\documentclass[letterpaper]{article} 
\usepackage{aaai24}  
\usepackage{times}  
\usepackage{helvet}  
\usepackage{courier}  
\usepackage[hyphens]{url}  
\usepackage{graphicx} 
\usepackage{natbib}  
\usepackage{caption} 
\usepackage{algorithm}
\usepackage{algorithmic}

\usepackage{newfloat}
\usepackage{listings}

\usepackage{enumitem}
\usepackage{newtxmath}

\usepackage{multirow}
\usepackage{makecell}
\usepackage{booktabs}

\newtheorem{requirement}{Requirement}

\DeclareCaptionStyle{ruled}{labelfont=normalfont,labelsep=colon,strut=off} 
\floatstyle{ruled}
\newfloat{listing}{tb}{lst}{}
\floatname{listing}{Listing}
\title{Zero-1-to-3: Domain-Level Zero-Shot Cognitive Diagnosis via One Batch of Early-Bird Students towards Three Diagnostic Objectives}
\author {
	Weibo Gao\textsuperscript{\rm 1,2},
	Qi Liu\textsuperscript{\rm 1,2}\thanks{Corresponding Author.}, 
	Hao Wang\textsuperscript{\rm 1,2}, 
	Linan Yue\textsuperscript{\rm 1,2}, 
	Haoyang Bi\textsuperscript{\rm 1,2}, 
	Yin Gu\textsuperscript{\rm 1,2}, \\
	Fangzhou Yao\textsuperscript{\rm 1,2},
	Zheng Zhang\textsuperscript{\rm 1,2},
	Xin Li\textsuperscript{\rm 1,3},
	Yuanjing He\textsuperscript{\rm 4}
}
\affiliations {
	\textsuperscript{\rm 1} University of Science and Technology of China \\
    \textsuperscript{\rm 2} Anhui Province Key Laboratory of Big Data Analysis and Application \& State Key Laboratory of Cognitive Intelligence \\
    \textsuperscript{\rm 3} Artificial Intelligence Research Institute, iFLYTEK Co., Ltd \\
	\textsuperscript{\rm 4} The Open University of China, China\\
	\{weibogao,lnyue,bhy0521,gy128,fangzhouyao,zhangzheng\}@mail.ustc.edu.cn, \{qiliuql,wanghao3,leexin\}@ustc.edu.cn, 
	\\ heyuanjing@ouchn.edu.cn
}

\usepackage{bibentry}

\begin{document}

\maketitle

\begin{abstract}
Cognitive diagnosis seeks to estimate the cognitive states of students by exploring their logged practice quiz data. It plays a pivotal role in personalized learning guidance within intelligent education systems.
In this paper, we focus on an important, practical, yet often underexplored task: domain-level zero-shot cognitive diagnosis (DZCD), which arises due to the absence of student practice logs in newly launched domains.
Recent cross-domain diagnostic models have been demonstrated to be a promising strategy for DZCD.
These methods primarily focus on how to transfer student states across domains.
However, they might inadvertently incorporate non-transferable information into student representations, thereby limiting the efficacy of knowledge transfer.
To tackle this, we propose \textit{Zero-1-to-3}, a domain-level zero-shot cognitive diagnosis framework via one batch of early-bird students towards three diagnostic objectives.
Our approach initiates with pre-training a diagnosis model with dual regularizers, which decouples student states into domain-shared and domain-specific parts.
The shared cognitive signals can be transferred to the target domain, enriching the cognitive priors for the new domain, which ensures the \textit{cognitive state propagation} objective.
Subsequently, we devise a strategy to generate simulated practice logs for cold-start students through analyzing the behavioral patterns from early-bird students, fulfilling the \textit{domain-adaption} goal.
Consequently, we refine the cognitive states of cold-start students as diagnostic outcomes via virtual data, aligning with the \textit{diagnosis-oriented} goal.
Finally, extensive experiments on six real-world datasets highlight the efficacy of our model for DZCD and its practical application in question recommendation.
The code is publicly available at \url{https://github.com/bigdata-ustc/Zero-1-to-3}.
\end{abstract}

\section{Introduction}
\begin{figure}[t]
	\centering
	\scalebox{0.36}
	{\includegraphics{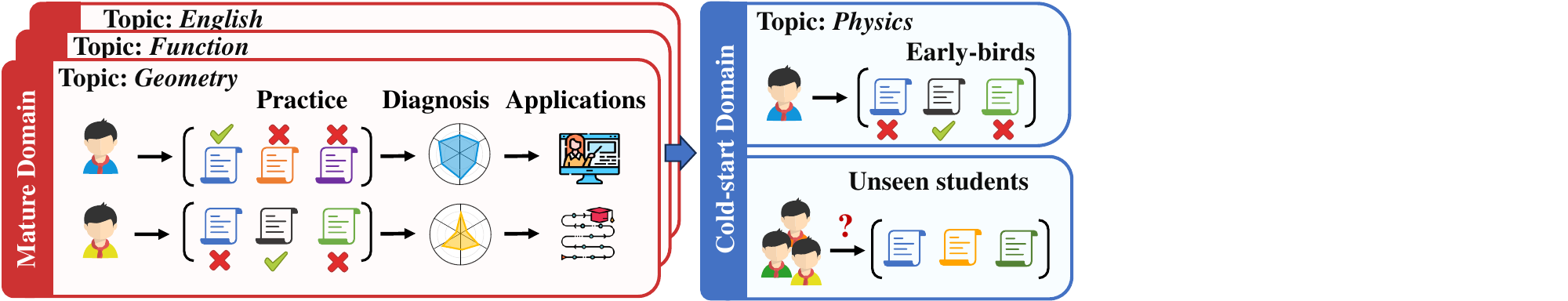}}
	\caption{Illustrative procedure of personalized learning in
intelligent education systems.}
	\label{fig:intro}
\end{figure}

Intelligent education systems offer access to learning resources and tailor-made services, contributing significantly to the burgeoning popularity of online learning. These platforms cover a broad range of learning topics. As shown in Figure~\ref{fig:intro}, each topic includes an extensive question bank, empowering students to independently select specific questions for targeted practice.
By analyzing student practice logs (e.g., correct or incorrect responses), their cognitive states (i.e., the proficiency on specific knowledge concepts) are estimated, which is referred to as the procedure of \textit{cognitive diagnosis}~\cite{wang2020neural}.
The diagnostic results can support further customized applications, such as question recommendation~\cite{liu2023meta} and adaptive testing~\cite{zhuang2023bounded}. As a result, cognitive diagnosis has garnered significant attention from both the ``AI for education" community and the general populace~\cite{nguyen2015effectiveness}.  

Previously, a number of cognitive diagnosis models (CDMs)~\cite{embretson2013item, reckase2009multidimensional, tsutsumi2021deep, wang2020neural, gao2021rcd, yao2023exploiting} have been developed to enhance diagnostic precision.
However, many of these models encounter challenges with the ``cold-start" problem.
This challenge arises when an online platform launches a novel learning topic with a fresh question bank (e.g., \textit{Physics} in Figure~\ref{fig:intro}). At the initial launch,  there exists only a limited collection of practice records from the early-bird learners, who form the first batch of students in this domain.
However, practice logs of unseen students remain unavailable for model training. Consequently, the diagnostic performance of traditional CDMs is often impaired as they only work in mature domains where student practice logs are available.
We call this task domain-level zero-shot cognitive diagnosis (DZCD).
DZCD is an important and practical task aiming to diagnose cognitive states of unseen students, for whom practice logs are blank in the new domain.

Recently, promising strategies have emerged to address the challenge of DZCD through cross-domain CDMs~\cite{gao2023leveraging}.
They primarily focus on the problem of \textbf{\textit{how to transfer}} student cognitive signals from well-established source domains to cold-start target domains.
The key point is to profile cognitive states of students based on their historical behaviors from some mature source domains and represent questions in the target domain with available features.
However, they tend to overlook a crucial but rarely touched concern: \textbf{\textit{what to transfer}}. This issue involves identifying valuable transferable student states from different topical domains, as not all cognitive signals can be transferred.
We argue that there exist two kinds of student cognitive preferences in each domain, namely \textit{domain-shared} and \textit{domain-specific}.
Let us consider \textit{Geometry} and \textit{Function} domains as an illustration. Cognitive signals shared across domains (e.g., basic skills) usually provide valuable clues for cross-domain transfer.
For example, students may perform similarly on simple questions in both domains because the knowledge examined in these straightforward questions is usually basic and general, e.g., \textit{Addition} and \textit{Subtraction}, which is suitable for cross-domain propagation.
In contrast, domain-specific states (e.g., high-level concept \textit{Cube} in \textit{Geometry} and concept \textit{Polynomials} in \textit{Function}) offer deeper insights within their respective domain, while they could be irrelevant or detrimental for other domains, which is commonly non-transferable.
For instance, understanding concept \textit{Cube} might not contribute significantly to mastering \textit{Polynomials}.
Unfortunately, previous models inevitably encode domain-specific information into student states intended for cross-domain transfer, which hampers transfer performance and can even the negative transfer problem~\cite{hu2018conet}.

In this paper, we aim to delve deeper into the problem of “what to transfer” to unlock the full potential of CDMs under DZCD, which is significant but challenging.
Ideally, a desirable solution should fulfill the following three objectives~\cite{gao2023leveraging}:
(1) \textit{diagnosis-oriented}: the model should proficiently diagnose student cognitive states in DZCD scenarios.
(2) \textit{cognitive signal propagation}: the model must effectively extract domain-shared states from source domains for the cross-domain propagation of student cognitive signals.
(3) \textit{domain-adaption}: for any new cold-start domain, the model is expected to be domain-adaptive, which needs to fully leverage available domain-specific cues in cold-start scenarios, e.g., the first batch of early-bird students in the domain.

Motivated by the above considerations, we propose \textit{Zero-1-to-3}, a domain-level \underline{zero}-shot cognitive diagnosis framework via \underline{one} batch of early-bird students \underline{to}wards \underline{three} diagnostic objectives. 
Specifically, our approach begins with the pre-training of a CDM across multiple source domains to establish an initial profile of students' states.
During this phase, we separate student profiles into domain-shared and domain-specific parts as input of the CDM. Meanwhile, two well-designed regularizers are placed to guide their optimizations.
After pre-training, the shared cognitive signals can be refined and can be transferred to the new domain to provide useful and broad experiences, which ensures the \textit{cognitive signal propagation} objective.
Next, our focus shifts to effectively adapting cold-start students whose practice logs are unavailable.
To achieve this, we first use domain-shared states as initial student embeddings in the new domain. Then, we devise a strategy to generate simulated practice logs for unseen students, to fine-tune the states for unseen students.
Here, we leverage the cognitive similarity between early-bird and unseen students to transfer practice patterns from the former to the latter, resulting in synthesized data. Notably, as the practice behaviors of early-bird students originate directly from the new domain, they offer distinct insights crucial for achieving the \textit{domain-adaption} goal.
After attaining warm-up states for cold-start students through fine-tuning using simulation logs, we can proceed with diagnostic predictions in the new domain characterizing our approach as \textit{diagnosis-oriented}.
Notably, it is important to highlight that Zero-1-to-3 framework, as a general framework, is applicable across a wide range of CDMs.

Finally, extensive experimental results on six real-world datasets not only prove that Zero-1-to-3 can effectively perform DZCD and outperform typical baselines, but also highlight a practical application in question recommendation.

\section{Related Work}
\textbf{Traditional Cognitive Diagnosis Models.}
Cognitive diagnosis has been well-researched for decades in educational psychology~\cite{bi2023beta, chen2023disentangling}.
It aims to profile the cognitive states of students by exploiting their response results (e.g., correct or wrong)~\cite{embretson2013item, tong2022incremental}. 
For instance, Item Response Theory (IRT)~\cite{embretson2013item} and Multidimensional IRT (MIRT)~\cite{reckase2009multidimensional} use unidimensional/multidimensional latent parameters indicating student ability and question difficulty, respectively, to predict student response on this question in a logistic way.
Deterministic Inputs, Noisy-And gate (DINA)~\cite{de2009dina}, NeuralCD~\cite{wang2020neural} and RCD~\cite{gao2021rcd} directly model student proficiency of specific knowledge concepts.

\noindent\textbf{Cross-domain Cognitive Diagnosis.}
Cross-domain cognitive diagnosis is proposed to address the DZCD issue which arises when an online education platform introduces new domains, resulting in unavailable practice logs for most students.
DZCD is a practical task, but research in this area is almost blank~\cite{gao2023leveraging}.
Existing studies~\cite{gao2023leveraging} on DZCD primarily concentrate on effectively transferring student cognitive signals from source domains to cold-start target domains through cross-domain modeling. The primary challenge is to construct student cognitive representations based on existing domains and utilize question attributes (e.g., textual contents~\cite{liu2019ekt, schmucker2022transferable} and question relational attributes~\cite{gao2023leveraging}) as intermediaries for the cross-domain transfer.
However, these methods might inadvertently incorporate non-transferable information into student representations limiting transfer performance.

\section{Preliminaries}
\subsection{Cognitive Diagnosis Model} \label{sec:cdm}
We first introduce the general form of cognitive diagnosis models (CDMs). We select the general form of CDMs proposed in TechCD~\cite{gao2023leveraging} as our framework, which consists of three types of basic elements: students, questions and knowledge concepts. The diagnosis process can be abstracted as modeling student-question-concept interactions through predicting student practice performances as follows:
\begin{equation}
    \label{eq:general_cd}
    \hat{y}_{u,v} = \mathcal{M}_{CD}(\boldsymbol{u},\boldsymbol{v},\mathbf{C}),\nonumber
\end{equation}
where $\boldsymbol{u}$ and $\boldsymbol{v}$ are the traits of student $u$ (e.g., cognitive states) and question $v$ (e.g., difficulty). $\mathbf{C}$ is the embedding matrix of all knowledge concepts.
$\mathcal{M}_{CD}(\cdot)$ is the diagnostic function to predict the response result $\hat{y}_{u,v}$ of student $u$ on question $v$ (correct or wrong).
For instance, $\mathcal{M}_{CD}(\cdot)$ is a logistic-like function for IRT/MIRT with uni-/multi-dimensional latent parameters of student ability and question difficulty, respectively, i.e., $\hat{y}_{u,v} = {\rm sigmoid}(\boldsymbol{u}-\boldsymbol{v})$, where $\boldsymbol{u}$ and $\boldsymbol{v}$ are enhanced by fusing concept features $\mathbf{C}$.
It is a multi-layer neural network $f(\cdot)$ in NeuralCD, i.e., $\hat{y}_{u,v} = f(\boldsymbol{q_{v}} \circ (\boldsymbol{p}_{u}-\boldsymbol{d}_{v}))$, where $\boldsymbol{p_{u}} \leftarrow f_{u}(\boldsymbol{u}, \mathbf{C})$ and $\boldsymbol{d_{v}} \leftarrow f_{v}(\boldsymbol{v}, \mathbf{C})$.
Two full connection layers $f_{u}(\cdot)$ and $f_{c}(\cdot)$ are used to fuse knowledge concepts into the student/question embeddings, respectively.
Each element ${p}_{u,c}$ of $\boldsymbol{p}_{u}$ denotes the mastery level on concept $c$ of student $u$. $\boldsymbol{q_{v}}$ masks unrelated concepts for question $v$ with element-wise product $\circ$, where ${q}_{v,c}=1$ if question $v$ associates concept $c$ and ${q}_{v,c}=0$ otherwise.

To ensure psychometric interpretability of prediction, CDMs should strictly follow the \textit{Monotonicity} assumption: the probability of correctly answering the question monotonically increases with student cognitive proficiency, i.e., $\partial \mathcal{M}_{CD} / \partial \boldsymbol{u} > 0$~\cite{tong2022incremental}.

\subsection{Problem Setup}
In the domain-level zero-shot cognitive diagnosis (DZCD) scenario, we consider $M$ source domains $\mathcal{S}^{1}, \mathcal{S}^{2}, \dots, \mathcal{S}^{M}$ and one target domain $\mathcal{T}$.
In the $k$-th source domain $\mathcal{S}^{k}$, suppose $\mathcal{U}_{\mathcal{S}^{k}}$, $\mathcal{V}_{\mathcal{S}^{k}}$ and $\mathcal{C}_{\mathcal{S}^{k}}$ are the sets of students, questions and concepts. The practice logs in this domain is depicted as $L_{\mathcal{S}^{k}}=\{(u^{k},v^{k}, y_{u,v}^{k})|y_{u,v}^{k} \in \{0,1\}, u^{k} \in \mathcal{U}_ {\mathcal{S}^{k}}, v^{k} \in \mathcal{V}_{\mathcal{S}^{k}}\}$, where $y_{u,v}^{k}=1$ represents student $u$ answers question $v$ correctly in source domain $\mathcal{S}^{k}$, and $y_{u,v}^{k}=0$ otherwise.
In the target domain $\mathcal{T}$, student set $\mathcal{U}_{\mathcal{T}}$ is split into two subsets: $\mathcal{U}_{\mathcal{T}^{(0)}}$ and $\mathcal{U}_{\mathcal{T}^{?}}$ denote the first batch of early-bird students and remaining unseen students, respectively. $\mathcal{V}_{\mathcal{T}}$ and $\mathcal{C}_{\mathcal{T}}$ are the sets of questions and concepts, respectively.
The practice records of $\mathcal{U}_{\mathcal{T}^{(0)}}$ is available, denoted as $L_{\mathcal{T}^{(0)}}=\{(u,v, y_{u,v})|y_{u,v} \in \{0,1\}, u \in \mathcal{U}_ {\mathcal{T}^{(0)}}, v \in \mathcal{V}_{\mathcal{T}}\}$.
The student set of the target domain is the subset of all the source domains, but question sets on each domain (including source and target domains) are totally different. Besides, let $|\mathcal{U}|$ denote the number of set $\mathcal{U}$, and $||\cdot||$ denote $\ell_{2}$ norm.

Based on the above setup, we aim to diagnose cognitive states for unseen cold-start students $\mathcal{U}_{\mathcal{T}^{?}}$ through fully exploiting available practice records (i.e., $L_{\mathcal{S}^{k}} \cup L_{\mathcal{T}^{(0)}}$) with student performance predictions.

\section{Methodology}
\begin{figure}[t]
	\centering
	\scalebox{0.47}	{\includegraphics{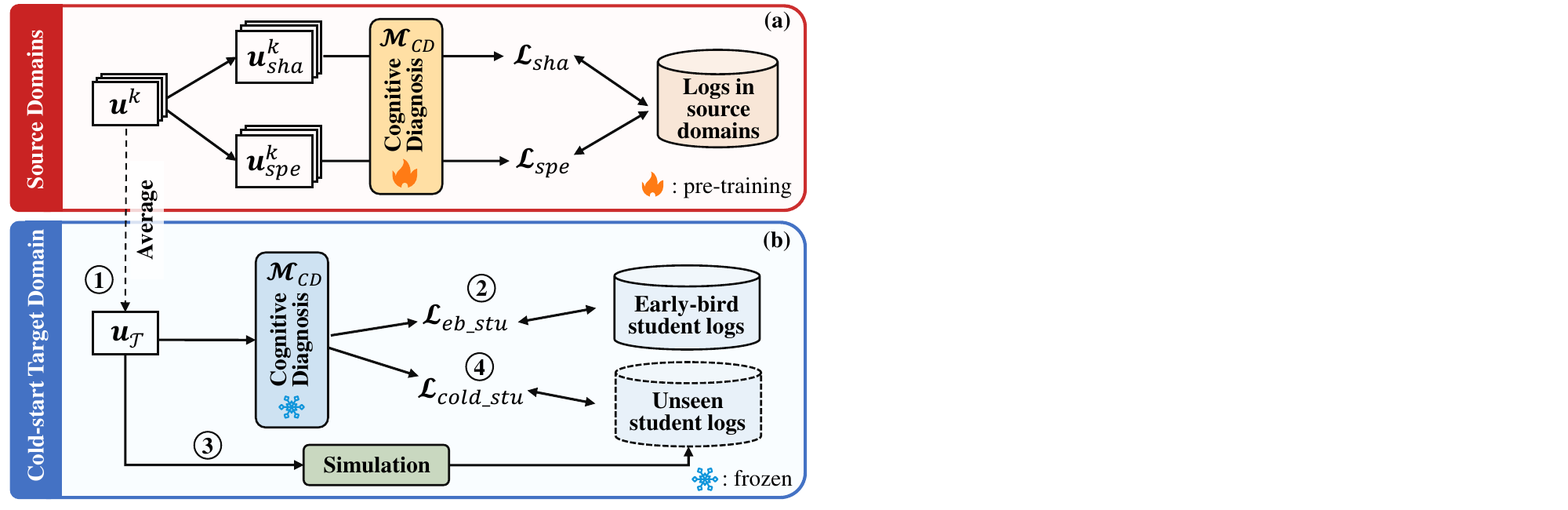}}
	\caption{The main framework of Zero-1-to-3. (a) shows the pre-training stage in source domains with cognitive state decoupling. (b) is the adaptive diagnosis stage in the new domain, where $\textcircled{1}\sim\textcircled{4}$ denote execution sequence. $\textcircled{1}$ is the initialization step, $\textcircled{2}$ refines early-bird student states, $\textcircled{3}$ simulates virtual logs for unseen students. The simulated logs are used to fine-tune cold-start students with $\textcircled{4}$.}
	\label{fig:framework}
\end{figure}

Our framework contains a pre-training process with cognitive state decoupling (Figure~\ref{fig:framework} (a)) and an adaptive diagnosis with fine-tuning via simulation logs (Figure~\ref{fig:framework} (b)). Next, we introduce them starting from an embedding layer.

\subsection{Embedding Layer}
This layer offers initialized representations for students, questions and knowledge concepts in each domain. Specifically, it contains several parameter matrices as student embeddings, i.e., $\mathbf{U}^{\mathcal{U}_{\mathcal{S}^{k}}} \in \mathbb{R}^{|\mathcal{U}_{\mathcal{S}^{k}}| \times F}$ and $\mathbf{U}^{\mathcal{U}_{\mathcal{T}}} \in \mathbb{R}^{|\mathcal{U}_{\mathcal{T}}| \times F}$ for students in source domain $\mathcal{S}^{k}$ and target domain $\mathcal{T}$, respectively, where $F$ is the dimensional size.
For the question, we adopt a pre-trained Bert~\cite{devlin2018bert} to encode its textual content as initial representations by averaging its word-level embeddings.
The question content-based embeddings in each domain includes $\mathbf{V}^{\mathcal{V}_{\mathcal{S}^{k}}} \in \mathbb{R}^{|\mathcal{V}_{\mathcal{S}^{k}}| \times F}$ and $\mathbf{V}^{\mathcal{V}_{\mathcal{T}}} \in \mathbb{R}^{|\mathcal{V}_{\mathcal{T}}| \times F}$. 
The concept embedding is obtained by averaging all relative question embeddings. We denote concept embeddings in source and target domains as $\mathbf{C}^{\mathcal{C}_{\mathcal{S}^{k}}} \in \mathbb{R}^{|\mathcal{C}_{\mathcal{S}^{k}}| \times F}$ and $\mathbf{C}^{\mathcal{C}_{\mathcal{T}}} \in \mathbb{R}^{|\mathcal{C}_{\mathcal{T}}| \times F}$, respectively. Note that the content-based encoder can cope with cold-start questions/concepts well by encoding their semantics~\cite{liu2019ekt}.
Our focus falls into student-side cross-domain transfer.

\subsection{Cognitive State Decoupling}
During this stage, a CDM is pre-trained across multiple source domains to establish an initial profile of students' states.
To ensure effective cross-domain transfer, the model should extract domain-shared cognitive representations from the input student embeddings within source domains.
However, prior cross-domain CDMs struggle to differentiate between general and specific signals, potentially hindering 
the effectiveness of cross-domain transfer.
Inspired by the success of decoupling learning in various fields~\cite{wang2023preference, liu2023homogeneous}, we propose to decouple student states into domain-shared and domain-specific parts.
The initial decoupled embeddings for students in each source domain can be obtained as follows:
\begin{equation}
    \label{eq:cross_entropy_loss}
    \boldsymbol{u}_{sha}^{k}=f_{sha}(\boldsymbol{u}^{k}), \; \boldsymbol{u}_{spe}^{k}=f_{spe}(\boldsymbol{u}^{k}),\nonumber
\end{equation}
where ${u}^{k}$ is $u$-th row of $\mathbf{U}^{\mathcal{U}_{\mathcal{S}^{k}}}$ denoting input trait for student $u$ in source domain $\mathcal{S}^{k}$ from embedding layer.
The input can be divided into:
$\boldsymbol{u}_{sha}^{k}$ containing domain-shared states while $\boldsymbol{u}_{spe}^{k}$ emphasizing domain-specific states. $f_{sha}(\cdot)$ and $f_{spe}(\cdot)$ are cross-domain shared full connection layers with the same input and output dimensions. 

Then, we devise two mathematical regularizers to optimize these two types of signals separately, inspired by~\cite{zhao2017physics, chen2018prerequisite, chen2022cerberus}.
Unlike prior decoupling methods, our insight is rooted in educational field-specific analysis, thoughtfully tailored for the DZCD task.

\textbf{Domain-shared states} reveal general cognitive preference across domains, providing a panoramic perspective for inferring student preferences. Leveraging these shared states can notably enhance the cross-domain transfer capability of CDMs. However, these broad clues cannot offer a fine peek into student preference in each individual domain as they are blocked by domain-specific information. Hence, domain-shared states should fulfill the following requirement:
\begin{requirement}
Suppose a CDM has obtained the domain-shared student representations from a domain $\mathcal{S}^{k}$. It may exhibit strong overall predictions across source domains. Besides, its performance within $\mathcal{S}^{k}$ is hindered by its absence of domain-specific information.
\end{requirement}

The above requirement can be approximatively expressed through the following regularization function:
\begin{equation}
    \label{eq:req_2}
    \begin{split}
    \mathcal{L}_{sha}
    &=\sum_{k=1}^{M}\Big(\mathbb{E}_{L_{\mathcal{S}^{i}}, i \in \{1,\cdots,M\}}
    \left|\left|y_{u,v}^{i} -\mathcal{M}_{CD}(\boldsymbol{u}_{sha}^{k},\boldsymbol{v}^{i},\boldsymbol{C}^{i})\right|\right| \\
    &-\sum_{(u^{k},v^{k},y_{u,v}^{k}) \in L_{\mathcal{S}^{k}}}\left|\left|y_{u,v}^{k} -\mathcal{M}_{CD}(\boldsymbol{u}_{sha}^{k},\boldsymbol{v}^{k},\boldsymbol{C}^{k}) \right|\right|\Big),\nonumber
    \end{split}
\end{equation}
where the first term minimizes the expectation of global prediction errors while the second term deliberately undermines prediction performance within a specific domain.

\textbf{Domain-specific states} contain student preferences for unique and specialized knowledge  within the relevant learning topic, often supplying richer in-domain insights. Logically, by harnessing these specific cues within a particular domain, CDMs can achieve enhanced diagnostic performance as these clues hold significant in-domain value.
However, this unique information might compromise the diagnostic accuracy across other domains, as it is typically irrelevant to unrelated domains.
Hereby, the domain-specific cognitive states should satisfy the following requirement: 
\begin{requirement}
Assume a CDM has obtained the domain-specific student states from $\mathcal{S}^{k}$. It probably leads to impressive predictive performance within domain $\mathcal{S}^{k}$. Conversely, prediction accuracy is expected to be suboptimal when the model is employed in other domains.
\end{requirement}

The above requirement can be further expressed through the minimization of the following regularization objective:
\begin{equation}
    \label{eq:req_1}
    \begin{split}
    \mathcal{L}_{spe}
    &=\sum_{k=1}^{M}\sum_{(u^{k},v^{k},y_{u,v}^{k}) \in L_{\mathcal{S}^{k}}}
    \Big(\left|\left|y_{u,v}^{k} -\mathcal{M}_{CD}(\boldsymbol{u}_{spe}^{k},\boldsymbol{v}^{k},\boldsymbol{C}^{k})\right|\right| \\
    &-\sum_{i \in \{1,\cdots,M\} \backslash \{k\}}\left|\left|y_{u,v}^{k} -\mathcal{M}_{CD}(\boldsymbol{u}_{spe}^{i},\boldsymbol{v}^{k},\boldsymbol{C}^{k}) \right|\right| \Big),\nonumber
    \end{split}
\end{equation}
where the first term narrows the gap between the actual performance $y_{u,v}^{k}$ and the prediction using the specific state of student $u^{k}$ on question $v^{k}$ from domain $\mathcal{S}^{k}$. The second term encourages misleading predictions by using domain-specific student states from other domains.

To summarize, the pre-training stage is directed by two regularizers for cognitive state decoupling as: $\mathcal{L}_{dec} = \mathcal{L}_{spe}+\mathcal{L}_{sha}$.
The shared cognitive signals from different domains can offer broad experiences that are useful when students encounter new areas, making it easier to transfer knowledge between domains.
Consequently, this process can preserve the \textit{cognitive signal propagation} objective.

\subsection{Domain-adaptive Cognitive Diagnosis}
After completing pre-training, our focus shifts towards performing domain-adaptive cognitive diagnosis for unseen students.
A simple yet effective solution is to directly use domain-shared states as initial student representations for DZCD, which has been used in previous studies~\cite{gao2023leveraging}. However, we contend that it is not domain-adaptive as it ignores in-domain considerations.
For this goal, we devise a strategy to generate simulated practice logs for unfamiliar students by fully utilizing available early-bird students.
Based on simulated practice logs, we can warm up student states by fine-tuning within the new domain.

Specifically, given a pre-trained CDM $\mathcal{M}_{CD}$ and decoupled embeddings of each student $u$, i.e., $\{\boldsymbol{u}_{sha}^{k}, \boldsymbol{u}_{spe}^{k}\}_{k=1}^{M}$, we freeze model parameters and initialize the state of each student $u \in \mathcal{U}_{\mathcal{T}}$ using his/her domain-shared representations from source domains inspired by the related techniques~\cite{YangZFJZ17, yue2023fedjudge} as follows:
\begin{equation}
    \label{eq:average}
    \begin{split}
    \boldsymbol{u}_{\mathcal{T}} = \frac{1}{M}\sum\nolimits_{k=1}^{M}\boldsymbol{u}_{sha}^{k},
    \end{split}
\end{equation}
where an average pooling is adopted to merge shared signals to obtain initial states $\boldsymbol{u}_{\mathcal{T}}$ of each student $u$ across $M$ source domains, since this operation can augment representations compared to ones in a single domain and smooth the biases across domains~\cite{wang2021hypersorec, zhu2023domain}. The domain-shared states are transferable, which can provide valuable priors.

Based on the setup, we utilize practice logs of the early-bird student $u \in \mathcal{U}_{\mathcal{T}^{(0)}}$ to refine their cognitive states, aiming to establish an initial understanding for the new domain. This process is optimized by minimizing the difference between prediction $\mathcal{M}_{CD}(\cdot)$ and actual response $y_{uv}$ as:
\begin{equation}
    \label{eq:fine_tune_1}
    \mathcal{L}_{eb\_stu}=\sum_{(u,v,y_{u,v}) \in L_{\mathcal{T}^{(0)}}}\left|\left|y_{u,v} -\mathcal{M}_{CD}(\boldsymbol{u}_{\mathcal{T}},\boldsymbol{v}_{\mathcal{T}},\boldsymbol{C}^{\mathcal{T}}) \right|\right|.\nonumber
\end{equation}

Next, we aim to generate simulated practice logs for remaining unseen students by exploiting in-domain clues from early-bird students.
Our basic assumption is that students with similar cognitive preferences are likely to achieve similar practice performance in a certain domain~\cite{long2022improving, 10.1145/3583780.3614781}. 
Thus, leverage the cognitive similarity between early-bird and unseen students to transfer practice patterns from the former to the latter, resulting in synthesized data.
With this in mind, we introduce a strategy to extract students' similarities from source domains as available clues in the target domain are very few.

In detail, for each early-bird student $u$, we first find a reference source domain $\mathcal{S}^{k}$ through the alignment of their refined embeddings in the target domain and domain-specific states originating from each source domain as follows:
\begin{equation}
    \label{eq:similar_domain}
    \mathcal{S}^{k} = \mathop{\arg\max}\limits_{k \in \{1,\cdots,M\}}\;{\rm sim}\left(\boldsymbol{u}_{\mathcal{T}},\boldsymbol{u}_{spe}^{k} \right),
\end{equation}
where ${\rm sim}(\cdot,\cdot)$ denotes the similarity function, for which we use cosine similarity.
$\mathcal{S}^{k}$ is selected as a reference domain as student $u$ has the most similar states between both $\mathcal{S}^{k}$ and the target domain.
Within $\mathcal{S}^{k}$, we compute similarity score between student $u$ and each unseen student $i$ by using their domain-specific representations as follows:
\begin{equation}
    \label{eq:similar_score}
    \begin{split}
       s_{u,i} \in \mathop{\rm sim}\limits_{i \in \mathcal{U}_{\mathcal{T}^{?}}}\left(\boldsymbol{u}_{spe}^{k},\boldsymbol{i}_{spe}^{k} \right),
    \end{split}
\end{equation}
where $\boldsymbol{u}_{spe}^{k}$ and $\boldsymbol{i}_{spe}^{k}$ are domain-specific states pre-trained in $\mathcal{S}^{k}$ of early-bird student $u$ and each unseen student $i$.
Then, we rank these unseen students via similarity scores and select top $p$ individuals to form a similar peer set for student $u$, denoted as $\mathcal{P}_{u}$.
Subsequently, for each selected unseen student in $\mathcal{P}_{u}$, denoted as $\hat{i}$, their virtual data can be approximatively generated by duplicating practice records of $u$ in the target domain as their performances are similar:
\begin{equation}
    \label{eq:simulate_log}
    \begin{split}
       L_{\mathcal{P}_{\hat{i}}} \leftarrow L_{\mathcal{T}_{u}} \subset L_{\mathcal{T}^{(0)}}.
    \end{split}
\end{equation}

By repeating the above processes (Eq.~(\ref{eq:similar_domain}) - (\ref{eq:simulate_log})) for each early-bird student, we can generate practice logs for parts of unseen students in the target domain. Note that the above simulation process might not cover every cold-start student, a concern linked to the parameter $p$. While a larger $p$ could involve more students, such an approach is not recommended due to the potential introduction of noise.
Based on simulation data, denoted as $L_{\mathcal{P}}$, we fine-tune the cognitive states of unseen students in the new domain as:
\begin{equation}
    \label{eq:fine_tune_2}
    \mathcal{L}_{cold\_stu}=\sum_{(u,v,y_{u,v}) \in L_{\mathcal{P}}}\left|\left|y_{u,v} -\mathcal{M}_{CD}(\boldsymbol{u}_{\mathcal{T}},\boldsymbol{v}_{\mathcal{T}},\boldsymbol{C}^{\mathcal{T}}) \right|\right|.
\end{equation}

Through optimization with the above loss, the cognitive embeddings of cold-start students can be augmented, ensuring our framework is \textit{diagnosis-oriented}.
Concurrently, this stage maximizes the utilization of domain-specific cues from the perspective of early-bird students, aligning with the \textit{domain-adaptation} objective.

\section{Experiments}
We conduct comprehensive experiments to address the following research questions:
\begin{itemize}[leftmargin=10pt]
\item \textbf{RQ1} How powerful is Zero-1-to-3 for the DZCD task?
\item \textbf{RQ2} How effective are the key components of the Zero-1-to-3 model?
\item \textbf{RQ3} Can the Zero-1-to-3 perform cognitive diagnosis well beyond the cold-start stage?
\item \textbf{RQ4} How to apply our framework to provide personalized learning guidance?
\end{itemize}

\subsection{Datasets} \label{sec:description}
We conduct experiments on six real-world datasets
i.e., Geometry, Function, Probability, Physics, Arithmetic and English, which are collected from the iFLYTEK Learning Machine\footnote{\url{https://xxj.xunfei.cn/}}.
All the datasets provide student practice records, question textual contents and question-concept correlations, where each question associates one knowledge concept.
For each dataset, we reserve only the first attempt of each question to ensure that student states are static following the~\cite{wang2020neural}.
Each dataset is treated as a domain.
We switch their roles, each acting as the cold-start target domain and leaving the other five as the source domains.
We split each source domain by randomly selecting two historical interactions from each student's logs for validation, with the remaining data serving as the training set, similar to the widely used \textit{leave-one-out} evaluation~\cite{rendle2009bpr}.
We randomly select some students (reported in section~\ref{sec:detail}) as early birds (the order in which student commence new domains does not influence one another) to introduce data diversity for each domain when acting as the target domain. 
Besides, to train the Oracle models (section~\ref{sec:baseline}), we also split the target domain's dataset into training (70\%), validation (10\%), and test sets (20\%). 
The basic statistics of datasets are listed in Table~\ref{tab:statistics}.


    
    
    


\begin{table}
\centering
\small
\renewcommand\arraystretch{1} 
\setlength{\tabcolsep}{0.5mm}{
{
\begin{tabular}{c|cccc}
    \toprule[1.0pt]
    \textbf{\quad Datasets\quad} & \;\#students\; & \;\#questions\; & \#concepts & \#logs \\
    \midrule
    Geometry & 15,283 & 2,299 & 251 & 127,570 \\
    Function & 15,404 & 2,172 & 201 & 121,512 \\
    Probability & 4,076 & 246 & 32 & 10,237 \\
    Physics & 13,369 & 2,699 & 552 & 146,326 \\
    Arithmetic & 14,073 & 1,828 & 200 & 86,699 \\
    English & 5,906 & 409 & 135 & 24,739 \\
    \midrule
    Total & 21,068 & 9,653 & 1,371 & \;517,083\; \\
    

    \bottomrule[1.0pt]
\end{tabular}}}
\caption{Some basic statistics of the datasets.}
\label{tab:statistics}
\end{table}

\begin{table*}[!h]
    \centering
    \setlength{\tabcolsep}{1mm}{
        {
            \begin{tabular}{c|ccccccccccccc}
            \toprule[1.0pt]
            \multicolumn{2}{c}{} &&
            \multicolumn{3}{c}{$\textbf{Geometry as Target}$} &&
            \multicolumn{3}{c}{$\textbf{Function as Target}$} &&
            \multicolumn{3}{c}{$\textbf{Probability as Target}$} \\
            
            \cmidrule{4-6}  \cmidrule{8-10}  \cmidrule{12-14}
            \multicolumn{2} {c} {\textbf{Methods}}&
            & \textbf{ACC\;$\uparrow$} & \textbf{AUC\;$\uparrow$} & \textbf{RMSE\;$\downarrow$} &
            & \textbf{ACC\;$\uparrow$} & \textbf{AUC\;$\uparrow$} & \textbf{RMSE\;$\downarrow$} &
            & \textbf{ACC\;$\uparrow$} & \textbf{AUC\;$\uparrow$} & \textbf{RMSE\;$\downarrow$} \\
            \midrule
            \multicolumn{2}{c}{Random} &
            & 49.91 & 49.83 & 57.80 & 
            & 49.97 & 50.04 & 57.76 & 
            & 50.17 & 50.96 & 57.60 
            \\
            \midrule
            \multirow{5}{*}{IRT}
            & \;Oracle\; &
            & $\;73.57^{*}$ & $\;79.86^{*}$ & $\;44.54^{*}$ & 
            & $\;74.16^{*}$ & $\;80.44^{*}$ & $\;42.71^{*}$ & 
            & $\;70.00^{*}$ & $\;72.41^{*}$ & $\;45.81^{*}$ 
            \\
            & NLP &
            & $\textbf{62.27}$ & $\underline{64.93}$ & $\textbf{48.62}$ & 
            & $\underline{55.38}$ & $\textbf{61.92}$ & $\textbf{49.15}$ & 
            & 58.51 & $\textbf{65.28}$ & $\textbf{48.73}$ 
            \\
            & GCN &
            & $\underline{60.49}$ & 62.92 & 51.62 & 
            & 54.23 & 56.67 & 52.12 &
            & 58.16 & 57.02 & 51.36 
            \\
            & Tech &
            & 59.96 & 62.50 & 49.86 & 
            & 52.45 & 59.58 & 55.13 & 
            & $\underline{65.22}$ & 54.24 & 54.58 
            \\
            & Ours &
            & 57.01 & $\textbf{64.96}$ & $\underline{49.51}$ & 
            & $\textbf{55.73}$ & $\underline{60.96}$ & $\underline{51.95}$ & 
            & $\textbf{65.37}$ & $\underline{58.30}$ & $\underline{50.47}$ 
            \\
            \midrule
            \multirow{5}{*}{MIRT}
            & Oracle &
            & $\;73.53^{*}$ & $\;80.57^{*}$ & $\;42.35^{*}$ & 
            & $\;74.03^{*}$ & $\;80.92^{*}$ & $\;41.82^{*}$ & 
            & $\;70.32^{*}$ & $\;72.76^{*}$ & $\;44.78^{*}$ 
            \\
            & NLP &
            & $\textbf{56.74}$& $\textbf{68.27}$ & ${52.15}$ & 
            & ${55.32}$ & $\underline{68.00}$ & 55.22 & 
            & $\underline{65.67}$ & $\underline{64.62}$ & $\underline{47.77}$ 
            \\
            & GCN &
            & 56.50 & 59.20 & 51.95 & 
            & $\underline{59.39}$ & 63.80 & $\underline{50.95}$ & 
            & 65.22 & 58.19 & 50.48 
            \\
            & Tech &
            & $\underline{56.66}$ & 63.02 & $\textbf{49.33}$ & 
            & 58.82 & 61.32 & 53.98 & 
            & 62.22 & 60.26 & 51.50 
            \\
            & Ours &
            & 51.48 & $\underline{65.98}$ & $\underline{50.56}$ & 
            & $\textbf{60.91}$ & $\textbf{68.01}$ & $\textbf{49.36}$ & 
            & $\textbf{68.27}$ & $\textbf{70.46}$ & $\textbf{45.60}$ 
            \\
            \midrule
            
            \multirow{5}{*}{\;NeuralCD\;}
            & Oracle &
            & $\;73.62^{*}$ & $\;81.10^{*}$ & $\;41.95^{*}$ & 
            & $\;74.27^{*}$ & $\;80.97^{*}$ & $\;41.99^{*}$ & 
            & $\;72.26^{*}$ & $\;75.25^{*}$ & $\;41.81^{*}$ 
            \\
            & NLP &
            & 58.66 & 58.16 & $\textbf{49.05}$ & 
            & $\underline{58.90}$ & $\textbf{57.85}$ & $\textbf{49.31}$ & 
            & 64.36 & 50.28 & $\underline{48.73}$ 
            \\
            & GCN &
            & 56.84 & 50.10 & $\underline{49.74}$ & 
            & 55.00 & 51.12 & 54.30 & 
            & 64.92 & 52.89 & 49.40 
            \\
            & Tech &
            & $\underline{60.20}$ & $\underline{60.23}$ & 53.44 & 
            & 58.33 & 52.83 & 52.63 & 
            & $\underline{65.22}$ & $\underline{53.53}$ & 57.23 
            \\
            & Ours &
            & $\textbf{61.48}$ & $\textbf{60.30}$ & 52.58 & 
            & $\textbf{59.78}$ & $\underline{55.11}$ & $\underline{52.04}$ & 
            & $\textbf{66.67}$ & $\textbf{67.43}$ & $\textbf{48.68}$ 
            \\
            \midrule
            \midrule
            \multicolumn{2}{c} {} &&
            \multicolumn{3}{c}{$\textbf{Physics as Target}$} &&
            \multicolumn{3}{c}{$\textbf{Arithmetic as Target}$} &&
            \multicolumn{3}{c}{$\textbf{English as Target}$} \\
            \cmidrule{4-6}  \cmidrule{8-10}  \cmidrule{12-14}
            \multicolumn{2} {c} {\textbf{Methods}}&
            & \textbf{ACC\;$\uparrow$} & \textbf{AUC\;$\uparrow$} & \textbf{RMSE\;$\downarrow$} &
            & \textbf{ACC\;$\uparrow$} & \textbf{AUC\;$\uparrow$} & \textbf{RMSE\;$\downarrow$} &
            & \textbf{ACC\;$\uparrow$} & \textbf{AUC\;$\uparrow$} & \textbf{RMSE\;$\downarrow$} \\
            \midrule
            \multicolumn{2}{c}{Random} &
            & 49.94 & 49.99 & 57.75 & 
            & 49.99 & 50.03 & 57.69 & 
            & 50.30 & 50.23 & 57.58 
            \\
            \midrule
            \multirow{5}{*}{IRT}
            & \;Oracle\; &
            & $\;70.46^{*}$ & $\;77.49^{*}$ & $\;43.91^{*}$ & 
            & $\;69.57^{*}$ & $\;69.14^{*}$ & $\;48.48^{*}$ & 
            & $\;74.37^{*}$ & $\;81.78^{*}$ & \;$40.96^{*}$ 
            \\
            & NLP &
            & $\underline{56.35}$ & $\textbf{59.63}$ & $\textbf{49.51}$ & 
            & $\textbf{60.66}$ & $\textbf{60.78}$ & $\underline{50.83}$ & 
            & $\textbf{64.03}$ & $\textbf{66.75}$ & $\textbf{48.37}$ 
            \\
            & GCN &
            & 54.32 & 52.08 & $\underline{52.03}$ & 
            & 54.04 & 51.12 & 51.03 & 
            & 54.34 & 50.90 & 55.13 
            \\
            & Tech &
            & 56.17 & $\underline{57.53}$ & 56.93 & 
            & 55.44 & 52.01 & 51.20 & 
            & $\underline{56.73}$ & 55.76 & 51.03 
            \\
            & Ours &
            & $\textbf{56.43}$ & 52.28 & 56.52 & 
            & $\underline{56.23}$ & $\underline{52.08}$ & $\textbf{50.72}$ & 
            & 54.46 & $\underline{58.28}$ & $\underline{50.07}$ 
            \\
            \midrule
            \multirow{5}{*}{MIRT}
            & Oracle &
            & $\;70.43^{*}$ & $\;77.45^{*}$ & $\;44.50^{*}$ & 
            & $\;74.60^{*}$ & $\;81.10^{*}$ & $\;42.68^{*}$ & 
            & $\;72.80^{*}$ & $\;81.01^{*}$ & $\;42.54^{*}$ 
            \\
            & NLP &
            & 54.32 & $\underline{61.18}$ & 54.96 & 
            & 60.01 & $\textbf{66.12}$ & 50.77 & 
            & 50.69 & $\underline{58.34}$ & 56.75 
            \\
            & GCN &
            & $\underline{54.35}$ & 58.17 & 57.52 & 
            & $\underline{60.53}$ & 60.02 & $\underline{50.60}$ & 
            & 50.47 & 51.15 & 54.41 
            \\
            & Tech &
            & 53.33 & 60.31 & $\underline{50.24}$ & 
            & 59.93 & 54.35 & 55.59 & 
            & $\underline{54.86}$ & 55.08 & $\underline{53.40}$ 
            \\
            & Ours &
            & $\textbf{56.71}$ & $\textbf{72.06}$ & $\textbf{48.71}$ & 
            & $\textbf{61.88}$ & $\underline{63.71}$ & $\textbf{48.52}$ & 
            & $\textbf{61.52}$ & $\textbf{66.17}$ & $\textbf{48.24}$ 
            \\
            \midrule
            
            \multirow{5}{*}{\;NeuralCD\;}
            & Oracle &
            & $\;70.22^{*}$ & $\;77.72^{*}$ & $\;43.94^{*}$ & 
            & $\;74.45^{*}$ & $\;81.80^{*}$ & $\;41.22^{*}$ & 
            & $\;72.64^{*}$  & $\;81.53^{*}$ & $\;41.98^{*}$ 
            \\
            & NLP &
            & $\underline{56.32}$ & $\underline{56.45}$ & 51.42 & 
            & $\underline{60.63}$ & 57.95 & $\textbf{49.43}$ & 
            & 50.47 & 50.48 & 51.12 
            \\
            & GCN & 
            & 54.21 & 54.54 & $\underline{50.26}$ & 
            & 60.44 & $\underline{58.40}$ & 52.68 & 
            & 50.47 & 52.15 & 50.51 
            \\
            & Tech & 
            & 54.23 & 52.18 & 50.81 & 
            & 56.99 & 52.40 & $\underline{49.57}$ & 
            & $\underline{57.93}$ & $\underline{56.12}$ & $\underline{50.10}$ 
            \\
            & Ours &
            & $\textbf{61.59}$ & $\textbf{68.08}$ & $\textbf{50.22}$ & 
            & $\textbf{60.89}$ & $\textbf{58.92}$ & 51.99 & 
            & $\textbf{60.87}$ & $\textbf{60.23}$ & $\textbf{48.96}$ 
            \\
            \bottomrule[1.0pt]	
            \end{tabular}
        }
    }
    \caption{Performance comparison (\%). The best zero-shot performance is highlighted in bold, and the runner-up is underlined. $\uparrow$ ($\downarrow$) means the higher (lower) score the better (worse) performance, the same as below. * indicates the oracle result.}
    \label{tab:prediction}
\end{table*} 

\subsection{Experimental Setup}
\subsubsection{Baselines} \label{sec:baseline}
To verify the effectiveness of our model, we apply our framework to three well-adopted CDMs, i.e., IRT~\cite{embretson2013item}, MIRT~\cite{reckase2009multidimensional} and NeuralCD~\cite{wang2020neural} (introduced in section~\ref{sec:cdm}). We call them Zero-CDM, e.g., Zero-IRT.
We select several baselines for comparison.
Among them, the Random and Oracle methods indicate the lower and upper bounds of performance.
For each baseline (excluding Random), we also select IRT, MIRT and NeuralCD as diagnostic functions.
\begin{itemize}[leftmargin=10pt]
    \item Random: The Random method predicts the students’ scores randomly from $Uniform(0,1)$.
    \item Oracle: The Oracle baseline is trained with logs from both source and target domains using the traditional CDM.
    \item NLP-based~\cite{liu2019ekt}:
    The NLP-based method uses learnable embeddings as student states in source domains and represents questions by encoding their texts. To implement it, we adopt Bert~\cite{devlin2018bert} as the textual encoder following the setup in~\cite{gao2023leveraging}.
    \item GCN-based and Tech-based~\cite{gao2023leveraging}: Both these methods use a knowledge graph~\cite{YangYBZZGXY23} linking each domain for transfer. The graph is constructed using a statistical method proposed in RCD~\cite{gao2021rcd}.
\end{itemize}

\subsubsection{Evaluation}
As cognitive states cannot be directly observed in practice, it is common to indirectly assess CDMs through predicting student performance on validation datasets.
To evaluate prediction performance, we adopt ACC and AUC, and RMSE as metrics from the perspectives of classification and regression, respectively, following previous works~\cite{gao2021rcd, zhang2023fairlisa}.

\subsubsection{Implementation Details} \label{sec:detail}
We set the dimensions of student and question vectors, i.e., parameter $F$, as the total number of knowledge concepts of each domain.
The dimensions of neural network layers are 512 and 256 for all NeuralCD-based models.
The number of early-bird students (($|\mathcal{U}_{\mathcal{T}^{(0)}}|$)) for each domain is set to be 0.01 of the number of students in the domain, and the number of unseen peer students ($p$) for the overall results of student performance prediction (RQ1) is 50.
We set the mini-batch size as 256 and select learning rate from \{0.001, 0.002, 0.02, 0.05\} for each model.
Each experiment is repeated five times under consistent conditions and the best score is reported.
Each model is implemented by PyTorch and optimized by Adam optimizer~\cite{kingma2014adam}.
All experiments are run on a Linux server with two 3.00GHz Intel Xeon Gold 5317 CPUs and one Tesla A100 GPU.
The code is publicly available at \url{https://github.com/bigdata-ustc/Zero-1-to-3}.


\begin{figure}[t]
	\centering
	\scalebox{0.65}
	{\includegraphics{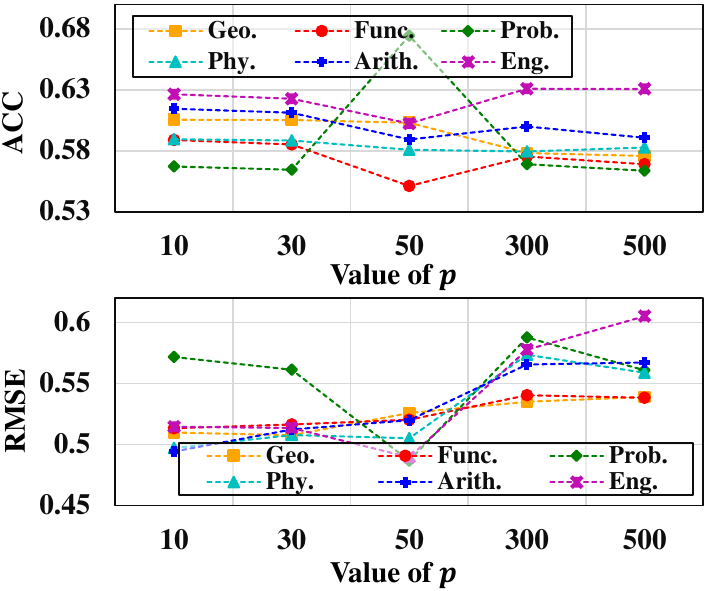}}
	\caption{Performance under different peer student numbers.}
	\label{fig:comprehensive_a}
\end{figure}

\begin{figure}[t]
	\centering
	\scalebox{0.74}
	{\includegraphics{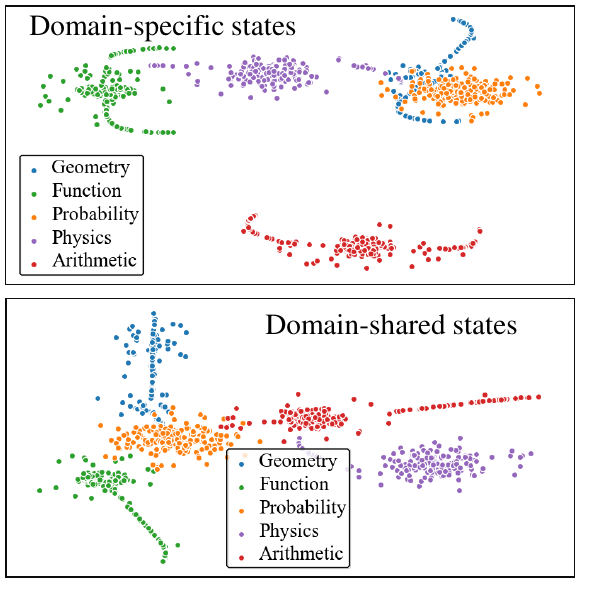}}
	\caption{T-SNE visualization of student states.}
	\label{fig:comprehensive_b}
\end{figure}

\begin{figure}[t]
	\centering
	\scalebox{0.65}
	{\includegraphics{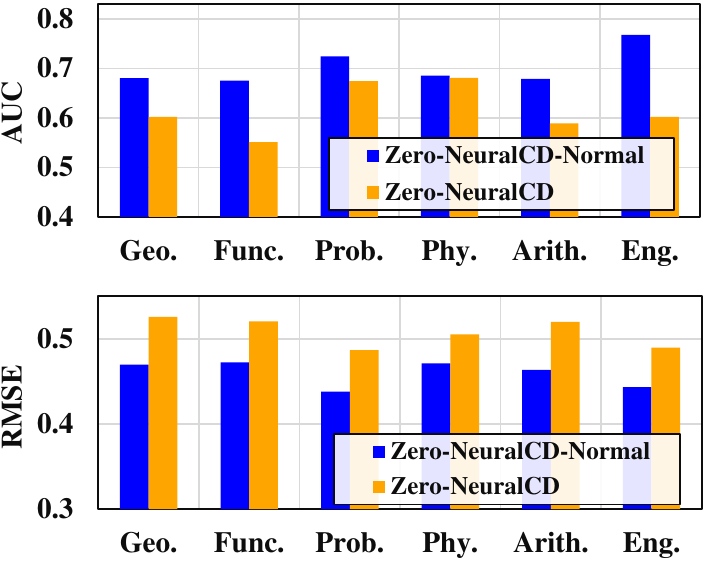}}
	\caption{Performance in normal diagnosis scenarios.}
	\label{fig:comprehensive_c}
\end{figure}
\subsection{Experimental Results}
\subsubsection{Student Performance Prediction (RQ1)}
We compare our model with several baselines on student performance predictions under DZCD.
We switch the role of each dataset acting as the target domain.
The overall performance is reported in Table~\ref{tab:prediction}.
We obtain that:
(1) For different diagnostic implementations (i.e., IRT, MIRT and NeuralCD as diagnostic models), our Zero-1-to-3 framework almost outperforms all baselines on each target domain, which indicates the diagnostic effectiveness of our solution under DZCD scenarios.
(2) The most significant distinction between Zero-1-to-3 and NLP methods lies in our consideration of decoupling student states and only transferring domain-shared states to address the ``what to transfer" issue. Our method outperforms NLP methods in most cases, underscoring the significance of adeptly capturing the domain-shared cognitive signals among students.
(3) Both GCN-based and Tech-based models employ a knowledge graph linking both source and target domains for domain-adaption by joint training. However, They cannot fully utilize domain-specific student logs. In contrast, Zero-1-to-3 outperforms these methods, which confirms its effectiveness.

In the following parts, we primarily present the experimental results of Zero-NeuralCD as the representative ones, since other diagnosis models can be abstracted as the special cases of NeuralCD~\cite{wang2020neural}.

\subsubsection{Detailed Analysis (RQ2)}
This section provides some in-depth analysis of how the key component in Zero-1-to-3 contributes to solving the challenges of DZCD.

\textbf{Exploration of Peer Student Number.}
The sampling number of early-bird students' peer students who are unseen ($p$) plays a crucial role in the transfer.
To study the impact of different numbers, we train Zero-NeuralCD under several settings and perform zero-shot student performance predictions on each dataset. 
We randomly sample the number of peer students from $\{10, 30, 50, 300, 500\}$.
Figure~\ref{fig:comprehensive_a} presents the results under different settings.
From the figure, we can see the diagnostic effect does not consistently improve with the increasing number of peer students during the simulation process, indicating that the violent simulation data will bring noise.
We are motivated to explore more suitable methods for matching peer students in future research.

\textbf{Visualization.}
This part visualizes student cognitive states to observe decoupling effects during pre-training.
Under the setting of ``English as Target", for the pre-trained Zero-NeuralCD (without fine-tuning), we randomly select 200 students and visualize their domain-specific/-shared states in each source domain using T-SNE~\cite{van2008visualizing} in Figure~\ref{fig:comprehensive_b}.
We can observe that domain-specific student states are more distant from each other compared to domain-shared student states, as they encompass more unique insights. Furthermore, domain-shared student states are not entirely blended, and we infer that this is because they still retain some personalized cognitive features that contribute to diagnostic performance.
These findings substantiate the efficacy of decoupling regularizers.

\subsection{Normal Diagnosis (RQ3)}
The above examples show Zero-1-to-3 can be successfully applied in the DZCD scenarios. A subsequent question is whether it can effectively perform cognitive diagnosis after the cold-start stage.
Thus, we conduct experiments by fine-tuning a Zero-NeuralCD (named Zero-NeuralCD-Normal) under the oracle setting where 70\% data in the target domain can be used for training.
The results under ``Geometry as Target" in Figure~\ref{fig:comprehensive_c} demonstrate a significant improvement in the performance of the model fine-tuned in a normal scenario compared to the model in a cold-start environment. This observation reveals that our model not only performs well in DZCD scenarios but also achieves satisfactory results in the subsequent stage, compared to Oracle models.

\subsection{Question Recommendation (RQ4)}
The above experiments have proved that Zero-1-to-3 can complete the DZCD task effectively. This part demonstrates one of the most popular diagnostic applications, i.e., question recommendation, that are in need of industrial practice.

We implement a simple yet effective recommendation strategy as an example to recommend $x$ questions for the student under DZCD.
Generally, a proper recommendation should not be too hard or easy to maintain students' enthusiasm when practicing~\cite{huang2019exploring}.
Thus, with a refined CDM, we first predict each student's performance on questions in the new domain via the $\mathcal{M}_{CD}(\cdot)$ in Eq.~(\ref{eq:fine_tune_2}).
All questions can be divided into two sets that answer correctly or not (i.e., positive or negative samples) according to prediction results.
Then, we sample $\frac{x}{2}$ questions from each type of sample, respectively, to yield the recommendation list.
\begin{table}[!t]
    \centering
    \centering
    \small
    \renewcommand\arraystretch{0.8} 
    \setlength{\tabcolsep}{0.8mm}{
    {
    \begin{tabular}{l|cccccc}
    \toprule[1.0pt]
    $\textbf{Geometry as Target}$ & $\textcircled{1}$ & $\textcircled{2}$ & $\textcircled{3}$ & $\textcircled{4}$ & $\textcircled{5}$ & $\textcircled{6}$ \\
    \midrule
    \text{Question id} & 3,213 & 200 & 1,032 & 2,122 & 3,013 & 32 \\
    \text{Mastery\;(\%)} & 40.07 & 32.00 & 48.63 & 26.30 & 54.33 & 41.02 \\
    \text{Difficulty\;(\%)} & 41.23 & 40.04 & 44.34 & 30.10 & 52.01 & 39.99 \\
    \text{True performance} & $\times$ & $\times$ & $\checkmark$ & $\times$ & $\checkmark$ & $\checkmark$ \\
    \bottomrule[1.0pt]
    \end{tabular}}}
    \caption{Question recommendation via Zero-NeuralCD.}
    \label{tab:recommendation}
\end{table}

Table~\ref{tab:recommendation} lists 6 recommended questions on target domain Geometry for a randomly selected student using a refined Zero-NeuralCD, the diagnosed student mastery levels and question difficulties of the associated concepts, and the student's true performance on the questions as recorded in the Geometry dataset. We can see the recommended questions are tailored to the student's proficiency, neither too easy nor too difficult. Some of them will challenge the student, while others will serve as ``gifts'' that can help increase his/her engagement. It confirms the application potential of Zero-1-to-3 in cold-start scenarios.

\section{Conclusion}
In this paper, we have proposed a general Zero-1-to-3 framework to tackle the real-world challenge of domain-level zero-shot cognitive diagnosis (DZCD).
The central obstacle in DZCD involves extracting shared information to facilitate cross-domain transfer, while simultaneously adapting to new domains.
To address this, we first pre-train a diagnosis model with dual regularizers that disentangle student states into domain-shared and domain-specific parts.
These shared cognitive signals can be transferred to the target domain, enriching cognitive priors for the new domain.
Subsequently, we develop a strategy to generate simulated practice logs for cold-start students in the target domain by using the behavioral patterns of early-bird students.
Consequently, the cognitive states of cold-start students can be adaptively refined using virtual data from the target domain, enabling the execution of DZCD.
Finally, extensive experiments highlight the efficacy and potential applicability of our framework.

\section*{Acknowledgments}
This research was partially supported by grants from the National Natural Science Foundation of China (Grant No. 62337001 and No. 62202443), the Anhui Provincial Natural Science Foundation (No. 2308085MG226), and the Fundamental Research Funds for the Central Universities.

\bibliography{aaai24}

\end{document}